\documentclass[11pt]{article}

\usepackage[final]{acl}

\usepackage{times}
\usepackage{latexsym}
\usepackage[T1]{fontenc}
\usepackage[utf8]{inputenc}
\usepackage{microtype}
\usepackage{graphicx}
\usepackage{amsmath}
\usepackage{amssymb}
\usepackage{booktabs}
\usepackage{xcolor}
\usepackage{tcolorbox}
\usepackage{listings}
\usepackage{algorithm}
\usepackage{hyperref}
\usepackage{cleveref}
\usepackage[noend]{algpseudocode}

\lstdefinestyle{ddoexample}{
  basicstyle=\small\ttfamily,
  breaklines=true,
  columns=fullflexible,
  keepspaces=true,
  showstringspaces=false,
  frame=single,
  framerule=0.3pt,
  rulecolor=\color{black},
  aboveskip=1pt,
  belowskip=2pt
}

\title{Direct Diversity Optimization for Diverse Successful Trajectories in Preference Post-Training}

\author{Junwon Ko \quad Dong-Jae Lee \quad Minchan Kwon \quad Sunghyun Baek \quad Junmo Kim \\
  School of Electrical Engineering, KAIST \\
  Daejeon, Republic of Korea \\
  \texttt{\{kojunewon,jhtwosun,kmc0207,baeksh\}@kaist.ac.kr} \\
  \texttt{junmo.kim@kaist.ac.kr}}

\begin{document}
\maketitle

\begin{abstract}
LLM agents for sequential decision tasks are often post-trained with trajectory-level outcome labels, but such labels provide little supervision for preserving multiple successful branches from the same decision state.
We study this problem as successful strategy coverage: how broadly a model realizes distinct successful strategies under a fixed rollout budget.
We present Direct Diversity Optimization (DDO), an offline post-training method that combines Divergence-Tree Collection (DTC) with the Reference-Relative Target-Odds Objective (RTO).\footnote{Code is available at \url{https://github.com/koguma00/direct_diverse_optimization}.}
DTC constructs state-aligned branch sets rooted at shared decision states, and RTO trains the model to match reference-relative targets over successful alternatives.
DDO achieves the strongest task success and successful strategy coverage among the compared post-training methods across BabyAI, BabaIsAI, and WebShop.
It also achieves the highest recovery rate after local action replacement and higher task success and coverage than successful-only imitation and decoding-time diversification controls.
\end{abstract}

\section{Introduction}
\label{sec:intro}

Recent work on Large Language Models (LLMs) commonly adapts pretrained models to downstream tasks through post-training~\citep{ouyang2022training,touvron2023llama,dubey2024llama,yang2025qwen3}.
The same approach is now applied to LLM agents in sequential decision tasks such as web navigation, embodied control, and tool use~\citep{zeng-etal-2024-agenttuning,song-etal-2024-trial,wei2025webagentr1}.
In these tasks, dense per-step supervision is rarely available, and post-training typically relies on trajectory-level success-or-failure feedback.
Under such coarse supervision, standard post-training tends to narrow the model's output distribution~\citep{kirk2024understanding,slocum2025diversepref,lanchantin2025diverse}.
In sequential decision tasks, this narrowing can cause the policy to collapse onto a single successful trajectory even when several viable alternatives remain.

\begin{figure}[t]
\centering
\includegraphics[width=\columnwidth]{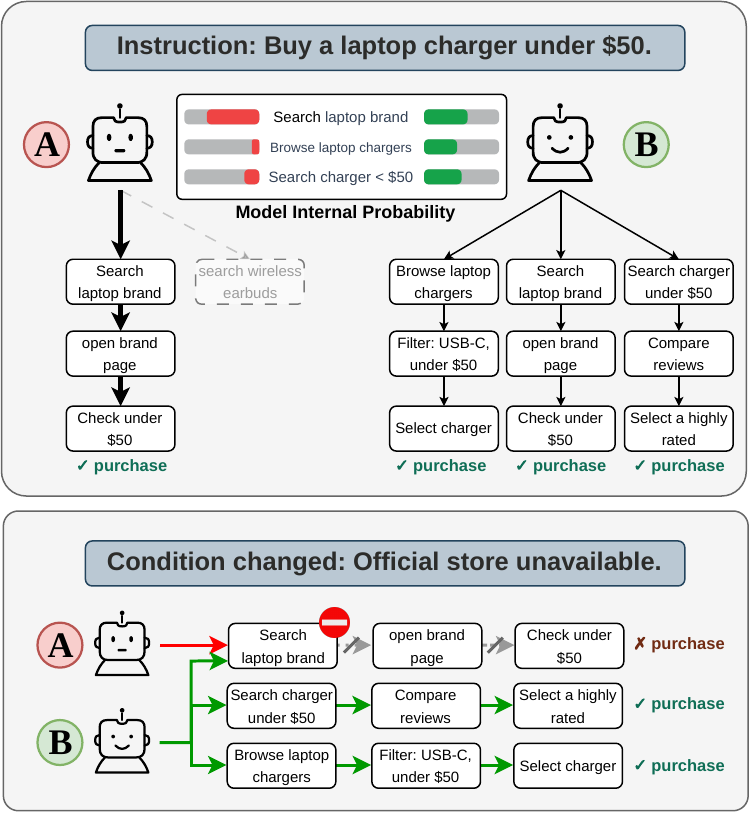}
\caption{A motivating online-shopping example: Model A, a standard post-trained model, completes the purchase through a single trajectory, while Model B retains multiple successful trajectories that can be used as fallbacks when conditions change.}
\label{fig:intro-agent-motivation}
\end{figure}

\Cref{fig:intro-agent-motivation} illustrates this collapse in an online-shopping task where an LLM agent must buy a laptop charger under a price constraint.
A standard post-trained model (Model A) commits to a single brand-page trajectory, whereas a model trained to retain multiple successful trajectories (Model B) can succeed through distinct search, filtering, and comparison strategies.
Both models succeed under normal conditions. When the brand-page trajectory is blocked, Model A has no fallback, whereas Model B retains an alternative path to success.

The contrast between Models A and B highlights a limitation of standard preference post-training: it lacks state-aligned supervision over alternative branches. Trajectory-level success labels identify a successful trajectory, but do not show the outcomes of alternative actions at the same decision state. Consequently, DPO-style training~\citep{rafailov2023direct} with success--failure trajectory preferences distinguishes successful from failed trajectories but provides no explicit target for allocating probability among multiple successful branches from the same state. The learned policy is therefore prone to placing most of its probability mass on one successful branch, leaving little for the remaining alternatives.

To address this limitation, we present Direct Diversity Optimization (DDO), an offline post-training method that couples a data-construction procedure with a reference-relative optimization objective.
Given a successful source trajectory, Divergence-Tree Collection (DTC) restores an intermediate decision state and constructs an outcome-labeled branch set containing the source and alternative branches.
Reference-Relative Target-Odds (RTO) replaces the single-winner preference target with a reference-relative target distribution over successful branches and fits the model to match its induced pairwise odds.
Together, DTC and RTO convert trajectory-level outcomes into preference supervision over multiple successful alternatives from the same decision state.

We evaluate DDO on BabyAI, BabaIsAI, and WebShop, which cover navigation, object interaction, multi-stage instruction following, rule manipulation, and web interaction.
Across the three benchmarks, DDO achieves the strongest joint performance on task success and successful strategy coverage among the compared post-training methods.
These gains also translate into more robust recovery after local action replacement.
DDO also achieves higher task success and coverage than successful-only imitation and decoding-time diversification controls.
Component analyses show complementary roles for DTC and RTO, whose combination yields the largest joint gains.

\section{Related Work}
\label{sec:related-work}

\subsection{Diversity- and tie-aware post-training.}
Preference- or reward-based post-training is now a standard mechanism for adapting LLMs with feedback~\citep{stiennon2020learning,ouyang2022training,bai2022training,rafailov2023direct}. 
However, it has also been shown to narrow the output distribution of trained models:
\citet{kirk2024understanding} and \citet{padmakumar2024writing} report that RLHF and DPO reduce output diversity relative to the base model, and \citet{slocum2025diversepref} show that KL-regularized preference learning can amplify dominant preferences.
To mitigate this collapse, recent approaches preserve multiple acceptable responses in the preference signal.
DivPO encourages diverse preferred responses through diversity-aware preference updates~\citep{lanchantin2025diverse}, while tie-aware DPO variants allow similarly preferred responses to be modeled as ties rather than forcing an arbitrary winner--loser direction~\citep{chen2026tiedpo}.
These approaches reduce the pressure to collapse acceptable outputs into a single mode, but they operate at the prompt-response or pairwise-comparison level.
Related goals have long been studied in reinforcement learning: maximum-entropy RL encourages stochastic high-return behavior, while skill-discovery and quality-diversity methods learn distinct skills or policy repertoires~\citep{haarnoja2018soft,eysenbach2018diversity,pugh2016quality,pierrot2022diversity}.
DDO targets offline preference post-training for sequential LLM agents, using state-aligned outcome-labeled branches and reference-relative targets to retain multiple observed successful branches.

\subsection{Post-training LLMs for sequential tasks.}
Preference-based post-training has been extended to sequential decision tasks using branching rollouts or step-level feedback.
Chain of Preference Optimization~\citep{zhang2024cpo} constructs preference data from alternative reasoning trajectories, while ETO~\citep{song-etal-2024-trial} contrasts successful and failed exploration trajectories at the trajectory level.
Other methods provide per-step signal: IPR~\citep{xiong2024watch} builds step-level contrastive pairs along expert trajectories, and AgentPRM~\citep{xi2025agentprm} trains a process reward model for step-wise progress.
Across these methods, the supervision signal primarily distinguishes better from worse trajectories or steps.
When multiple branches from the same state are successful, they are not explicitly modeled as a set of valid alternatives; they are typically paired against failures, folded into independent comparisons, or scored by a progress signal.

DDO uses the labeled branch set at a decision state as the supervision unit.
It preserves multiple successful branches as valid positives while distinguishing them from failed branches, without forcing an arbitrary ranking among the successful ones.
This couples diversity preservation directly with task success.

\begin{figure*}[t]
\centering
\includegraphics[width=0.96\textwidth]{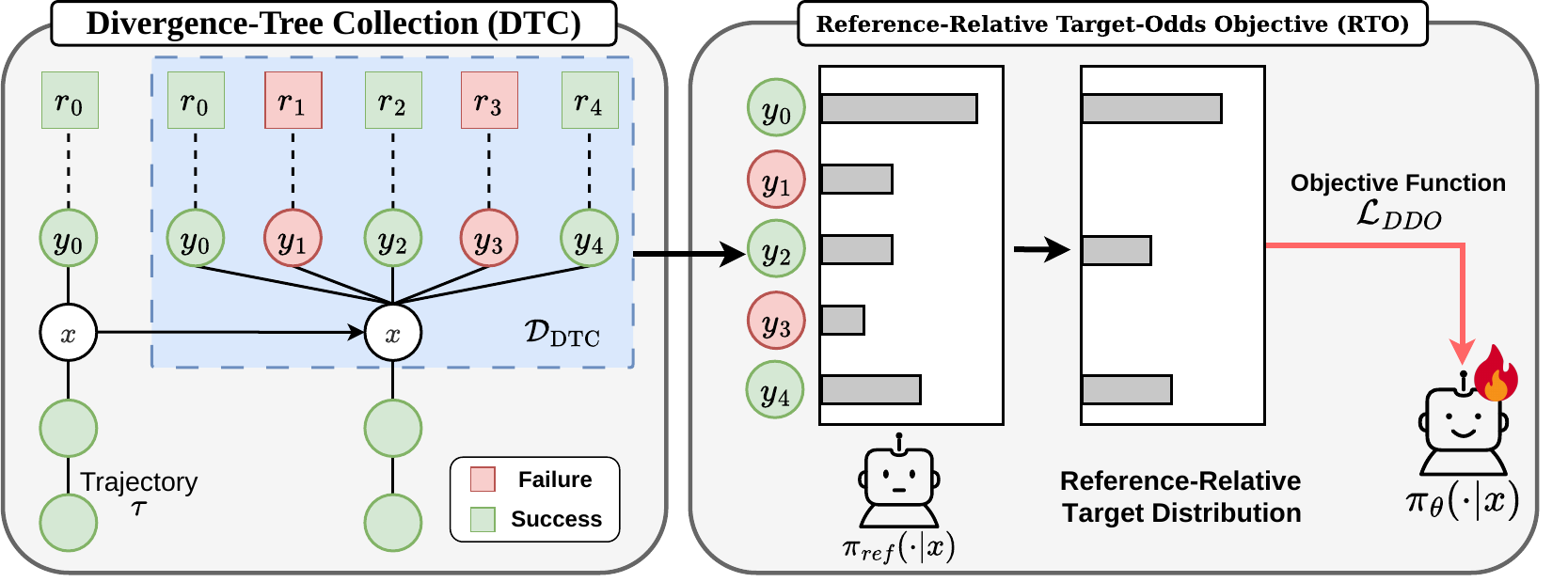}
\caption{DDO pipeline. DTC constructs state-aligned branch sets, and RTO defines a reference-relative target distribution whose induced pairwise targets are optimized with a soft logistic objective.}
\label{fig:method-overview}
\end{figure*}

\section{Method}
\label{sec:method}

We introduce Direct Diversity Optimization (DDO), an offline post-training method for retaining multiple successful strategies in LLM agents trained from trajectory-level outcome labels.
DDO couples Divergence-Tree Collection (DTC; \Cref{sec:dtc}), which constructs state-aligned branch sets with rollout outcome labels, with the Reference-Relative Target-Odds Objective (RTO; \Cref{sec:rto}), which fits the model to reference-relative targets over successful branches.
This coupling transforms trajectory-level outcome supervision into same-state branch comparisons and a target distribution over successful alternatives.
\Cref{fig:method-overview} summarizes the pipeline.

\subsection{Problem Setting and Notations}
\label{sec:m-not}

We address post-training of LLM agents on sequential decision tasks with textual observations and executable actions.
At each step \(t\), the agent observes a textual state \(x_t\) that includes the current observation and the recent interaction history, and produces an output \(y_t\); the environment executes the action parsed from \(y_t\) to produce the next state.
A trajectory \(\tau=(x_0,y_0,\ldots,y_{T-1},x_T)\) is judged only at termination, receiving a binary result \(R(\tau)\in\{0,1\}\) provided by the environment.
This trajectory-level signal is the sole supervision; no per-step rewards or learned reward model are assumed.
We write \(\pi_\theta(y\mid x)\) for the trainable model and \(\pi_{\mathrm{ref}}(y\mid x)\) for the frozen reference model.

\subsection{Divergence-Tree Collection}
\label{sec:dtc}
DTC takes a successful trajectory and produces, at selected decision steps, branch sets rooted at a shared state with rollout outcome labels.
For a source trajectory \(\tau\), the candidate divergence steps are the interior decision points \(1,\ldots,|\tau|-1\).
Under a fixed collection budget, DTC selects divergence steps to cover different depths along the source, so that branch roots are spread along the trajectory rather than concentrated near its initial prefix.
At each selected divergence step, DTC restores the corresponding decision state as the shared state \(x\) (\Cref{alg:dtc}).
It keeps the source output as one branch and queries an expert model for alternative outputs whose parsed actions differ from the source action.
Each retained output is executed to termination by the expert model and receives a binary success label.
If the state cannot be restored or no distinct valid alternative is obtained, DTC skips that state.

\begin{algorithm}[t]
\small
\caption{Divergence-Tree Collection (DTC). Implementation details of \textsc{Select}, \textsc{Restore}, \textsc{Alt}, and \textsc{Reject} are provided in Appendix~\ref{app:dtc}.}
\label{alg:dtc}
\begin{algorithmic}[1]
\Statex \textbf{Input:} successful sources \(\mathcal T^+\), expert model \(E\), budgets \(K_d,K_a\)
\Statex \textbf{Output:} DTC records \(\mathcal D_{\mathrm{DTC}}\)
\State \(\mathcal D_{\mathrm{DTC}}\leftarrow\emptyset\)
\For{\(\tau\in\mathcal T^+\)}
    \State \(D\leftarrow\textsc{Select}(\{1,\ldots,|\tau|-1\},K_d)\)
    \For{\(d\in D\)}
        \State \(x\leftarrow\textsc{Restore}(\tau,d)\)
        \If{\(x=\bot\)}
            \State \textbf{continue}
        \EndIf
        \State \(B\leftarrow\{j_{\mathrm{src}}\}\), \(y_{j_{\mathrm{src}}}\leftarrow y_d^\tau\), \(r_{j_{\mathrm{src}}}\leftarrow 1\)
        \State \(\mathcal Y\leftarrow\textsc{Alt}(E,x,y_d^\tau,K_a)\)
        \For{\(y\in\mathcal Y\)}
            \If{\(\textsc{Reject}(y,B)\)}
                \State \textbf{continue}
            \EndIf
            \State \(r\leftarrow R(\textsc{Rollout}(E,x,y))\)
            \State add new \(j\) with \((y_j,r_j)\leftarrow(y,r)\) to \(B\)
        \EndFor
        \If{\(B\neq\{j_{\mathrm{src}}\}\)}
            \State add \(\textsc{Record}(x,B)\) to \(\mathcal D_{\mathrm{DTC}}\)
        \EndIf
    \EndFor
\EndFor
\State \Return \(\mathcal D_{\mathrm{DTC}}\)
\end{algorithmic}
\end{algorithm}

Each collected branch set is represented as one DTC record:
\begin{equation}
C=(x,B,\{(y_j,r_j)\}_{j\in B}).
\label{eq:C}
\end{equation}
Here, \(x\) is the shared decision state, \(B\) is a finite set of branch indices, \(y_j\) is the LLM output of branch \(j\), and \(r_j\in\{0,1\}\) is the binary result obtained after executing that branch.
For each record we partition \(B\) into successful branches \(S^+=\{j:r_j=1\}\) and non-successful branches \(S^0=\{j:r_j=0\}\).
RTO uses \(S^+\) for target-odds estimation and \(S^0\) for boundary comparisons between successful and non-successful observed branches.

\subsection{Reference-Relative Target-Odds Objective}
\label{sec:rto}

Pairwise preference objectives that contrast a single winner against a single loser with hard binary labels risk collapsing \(\pi_\theta\) onto a single successful strategy.
The Reference-Relative Target-Odds (RTO) objective addresses this by aligning the log-odds \(\pi_\theta\) assigns to pairs of successful branches with the log-odds prescribed by a fixed distribution \(q\) over \(S^+\).
For each record \(C=(x,B,\{(y_j,r_j)\}_{j\in B})\), RTO constructs \(q\) from \(\pi_{\mathrm{ref}}\) and matches the two log-odds in reference-relative form, aligning \(\pi_\theta\)'s shift over \(\pi_{\mathrm{ref}}\) with \(q\)'s shift over \(q_{\mathrm{ref}}\).
Two pair types are handled asymmetrically: success--success pairs use \(q\) to set the desired log-odds within \(S^+\), and success--failure pairs across \(S^+\) and \(S^0\) use task success labels to order successful branches above failed ones.

\paragraph{Branch distribution.}
On \(S^+\), define the renormalized reference distribution \(q_{\mathrm{ref}}\) as
\begin{equation}
q_{\mathrm{ref}}(j) = \frac{\pi_{\mathrm{ref}}(y_j \mid x)}{\sum_{h \in S^+} \pi_{\mathrm{ref}}(y_h \mid x)}.
\label{eq:qref}
\end{equation}
The distribution \(q\) is defined as
\begin{equation}
q(j) = \frac{q_{\mathrm{ref}}(j)^\alpha}{\sum_{h \in S^+} q_{\mathrm{ref}}(h)^\alpha},
\label{eq:q}
\end{equation}
where \(\alpha \in [0, 1]\) controls how strongly \(q\) preserves \(\pi_{\mathrm{ref}}\)'s relative weighting among successful branches: smaller \(\alpha\) flattens these differences, larger \(\alpha\) keeps them.
At \(\alpha=0\), \(q\) is uniform on \(S^+\), providing maximal within-success diversity pressure; at \(\alpha=1\), \(q=q_{\mathrm{ref}}\) and the success--success targets reduce to preserving \(\pi_{\mathrm{ref}}\)'s relative odds among successful branches.
Both \(q\) and \(q_{\mathrm{ref}}\) are computed once from \(\pi_{\mathrm{ref}}\) and held fixed during training.
We use \(\alpha=0.5\) unless otherwise stated.

\paragraph{Distribution margin.}
For an ordered pair \(u, v \in S^+\) of successful branches, the target--reference log-odds margin is given by
\begin{equation}
m^\star = \log\frac{q(u)}{q(v)} - \log\frac{q_{\mathrm{ref}}(u)}{q_{\mathrm{ref}}(v)}.
\label{eq:ms}
\end{equation}
The margin \(m^\star\) is therefore the reference-relative log-odds shift needed for \(\pi_\theta\) to match \(q\) at \((u,v)\).
The corresponding target is \(p^\star = \sigma(\beta\, m^\star)\) with logistic scale \(\beta > 0\); for a success--failure pair with \(u \in S^+\) and \(v \in S^0\), \(q\) is undefined on \(v\) and we set \(p^\star = 1\) directly, so task success labels enter the objective only through these pairs.

\paragraph{Model margin.}
For \(\pi_\theta\), the analogous margin is defined in the same form so that it can be directly compared to \(m^\star\):
\begin{equation}
m_\theta = \log\frac{\pi_\theta(y_u \mid x)}{\pi_\theta(y_v \mid x)} - \log\frac{\pi_{\mathrm{ref}}(y_u \mid x)}{\pi_{\mathrm{ref}}(y_v \mid x)}.
\label{eq:m}
\end{equation}
Aligning \(m_\theta\) with \(m^\star\) is equivalent to matching \(\pi_\theta\)'s log-odds at \((u, v)\) to \(q\)'s, which is what the loss enforces.

\paragraph{Objective function.}
We train \(\pi_\theta\) by minimizing
\begin{equation}
\mathcal L_{\mathrm{DDO}}(\theta)
= \mathbb E\bigl[ D_{\mathrm{KL}}\bigl(P^\star \,\|\, P_\theta\bigr) \bigr],
\label{eq:L}
\end{equation}
where \(P^\star = \mathrm{Bern}(p^\star)\) and \(P_\theta = \mathrm{Bern}(p_\theta)\) are Bernoulli distributions with parameters \(p^\star\) and \(p_\theta = \sigma(\beta\, m_\theta)\).
The expectation is estimated over materialized pairs: each unordered success--success pair is included once, success--failure pairs are oriented toward the successful branch, and state-normalized weights prevent larger branch sets from dominating.
Up to a \(\theta\)-independent constant, this is equivalent to soft binary cross-entropy with target \(p^\star\) and prediction \(p_\theta\).
The success--success terms match the target odds \(q(u)/q(v)\) between successful branches, while success--failure terms set \(p^\star=1\) and reduce to the DPO-style loss \(-\log\sigma(\beta m_\theta)\), with \(u\in S^+\) and \(v\in S^0\).

\section{Experimental Setup}

\subsection{Benchmarks}

We evaluate DDO on three benchmarks that span complementary forms of sequential text-agent behavior.
BabyAI~\citep{chevalierboisvert2019babyai} probes controlled navigation, object interaction, and multi-stage instruction following.
BabaIsAI~\citep{cloos2024baba} targets rule manipulation, where the agent must change or exploit environment dynamics.
For BabyAI and BabaIsAI, we use the BALROG text-action implementations~\citep{paglieri2025balrog}, which provide standardized textual observations and executable action interfaces.
WebShop~\citep{yao2022webshop} examines web-style shopping behavior involving search, product inspection, option selection, and purchase decisions.
Details of each benchmark are provided in Appendix~\ref{app:benchmark-details}.

\subsection{Baselines and Comparison Protocol}

We compare DDO against DPO and diversity-aware post-training baselines.
DPO is the standard Direct Preference Optimization baseline~\citep{rafailov2023direct}.
DivFreq and DivProb adapt DivPO to state-aligned branch data by treating each shared state as a prompt, its executable branch outputs as candidate responses, and terminal rollout outcomes as quality labels~\citep{lanchantin2025diverse}.
They retain DivPO's frequency- and probability-based pair-selection criteria, respectively.
TieDPO-RK and TieDPO-Dav are tie-aware DPO baselines based on the Rao-Kupper and Davidson variants of \citet{chen2026tiedpo}.

We use \mbox{Qwen3-1.7B} as the target model and \mbox{Qwen3.5-122B-A10B-FP8} as the expert model for DTC collection~\citep{yang2025qwen3,qwen2026qwen35}.
Base denotes the unadapted target model before task SFT or preference optimization.
Reference denotes the frozen task-SFT model, fine-tuned from the base on the successful trajectories that also serve as DTC sources.
It is used both as the initialization for all post-training methods and as the reference model for reference-relative objectives.

For fair comparison, all post-training methods within a benchmark use the same DTC-collected branch sets, so the comparison isolates objective-side differences while holding the data side fixed.
The DTC ablation in \Cref{sec:dtc-results} separately varies the data side by replacing this resource with comparisons formed without DTC, allowing us to test the contribution of the collection procedure itself.
Training and evaluation otherwise follow the same task budgets and decoding protocols within each benchmark.
Full experimental details are provided in Appendix~\ref{app:expdetails}.

\subsection{Metrics}
\label{sec:metrics}

We evaluate each model along two axes: task success and successful strategy coverage under a fixed rollout budget.
A rollout is \emph{valid} if it terminates without system failures or missing model outputs.
Success rate is the fraction of valid rollouts that solve the task. 
BabyAI and BabaIsAI count a rollout as successful when the benchmark progression reaches \(1.0\).
WebShop returns a graded purchase score rather than a binary success label; we treat scores of at least \(0.9\) as successful, using the same fixed threshold across all methods.

To measure successful strategy coverage, we group successful rollouts into trajectory classes using a per-benchmark equivalence relation that abstracts surface variation in the action sequence (see Appendix~\ref{app:benchmark-details} for the per-benchmark rules).
Two successful rollouts belong to the same class when they are equivalent under this relation.
Within each benchmark, we apply the same equivalence relation to the rollouts from all evaluated methods.
We use effective strategy diversity (ESD) and entropy effective strategy diversity (H-ESD) to summarize the resulting classes.
Both metrics are success-restricted and budget-normalized: they are computed from successful trajectory classes but normalized by the total number of valid rollouts.

For each evaluation item \(i\), let \(K_i\) be the number of valid rollouts and \(U_i\) the number of unique successful trajectory classes.
Let \(H_i\) be the entropy of the empirical distribution over these classes:
\[
\mathrm{ESD}(i)=\frac{U_i}{K_i},
\qquad
\mathrm{H\text{-}ESD}(i)=\frac{2^{H_i}}{K_i}.
\]
ESD counts how many distinct successful trajectory classes are observed under the rollout budget, while H-ESD is smaller when successful rollouts concentrate on a few repeated trajectory classes.
Normalizing by \(K_i\), rather than by the number of successful rollouts, makes the metric reflect finite-budget successful strategy coverage: a model receives a high score only when it both solves the task and produces distinct successful trajectory classes.
Reported scores are uniform means of the item-level ESD and H-ESD over the evaluation items in each task.
Both metrics are zero when no successful rollout is observed.

\section{Results}

We first evaluate DTC as a data-construction procedure by measuring collection-time successful strategy coverage and downstream performance with and without DTC.
We then fix DTC and compare post-training objectives across the three benchmarks, isolating RTO under the same state-aligned branch supervision.

\subsection{DTC Improves Branch Supervision}
\label{sec:dtc-results}

\begin{figure}[t]
\centering
\includegraphics[width=\columnwidth]{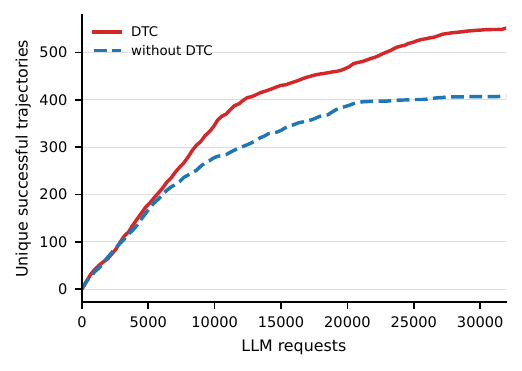}
\caption{Effect of DTC on BabyAI and BabaIsAI. Each curve reports the task-averaged number of unique successful trajectories at a matched per-task request budget, under benchmark-specific trajectory normalization.}
\label{fig:dtc-trajectory-budget}
\end{figure}

To isolate the data side of DDO, we run DPO and DDO with and without DTC.
DTC converts trajectory-level outcome labels into same-state branch supervision. For the collection-efficiency comparison, Without DTC independently samples full trajectories with the same expert under the matched request budget. For downstream training, settings without DTC form same-task comparisons from separately sampled rollouts matched to the training comparison exposure.

\Cref{fig:dtc-trajectory-budget} shows that DTC discovers more unique successful trajectories under the same request budget on both BabyAI and BabaIsAI, thereby increasing collection-time successful strategy coverage before post-training begins.

\begin{figure}[t]
\centering
\includegraphics[width=\columnwidth]{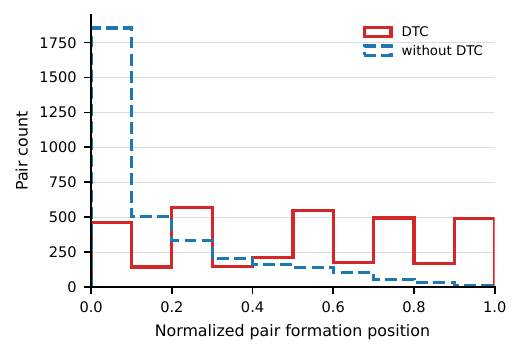}
\caption{Distribution of normalized pair-formation positions along source trajectories with and without DTC. Positions are normalized by trajectory length.}
\label{fig:dtc-normalized-position}
\end{figure}

DTC pairs are distributed across the trajectory, with \(56.4\%\) formed at or beyond the midpoint.
Pairs without DTC are concentrated near the initial prefix: \(79.3\%\) occur within the first \(20\%\) of the source trajectory.
This structural difference matters for sequential strategies, because early, middle, and late decisions can play different roles in reaching success.
DTC therefore provides branch-level supervision across a broader range of decision depths.

\begin{table}[b]
\centering
\scriptsize
\def\metrichead#1{#1\,\(\uparrow\)}
\setlength{\tabcolsep}{2pt} %
\begin{tabular*}{\columnwidth}{@{\extracolsep{\fill}}lc@{\hspace{0.35em}}cccccc@{}}
\toprule
\multicolumn{1}{@{}l@{\hspace{0.25em}}}{Method} & DTC &
\multicolumn{3}{c}{BabyAI} &
\multicolumn{3}{c}{BabaIsAI} \\
\cmidrule(lr){3-5} \cmidrule(lr){6-8}
 & & \metrichead{Succ.} & \metrichead{H-ESD} & \metrichead{ESD} & \metrichead{Succ.} & \metrichead{H-ESD} & \metrichead{ESD} \\
\midrule
DPO & \(\times\) & 0.76 & 0.38 & 0.44 & 0.78 & 0.26 & 0.30 \\
DPO & \(\checkmark\) & 0.84 & 0.36 & 0.43 & 0.94 & 0.31 & 0.38 \\
DDO & \(\times\) & 0.87 & 0.39 & 0.46 & 0.83 & 0.28 & 0.35 \\
DDO & \(\checkmark\) & \textbf{0.92} & \textbf{0.45} & \textbf{0.52} & \textbf{0.99} & \textbf{0.38} & \textbf{0.47} \\
\bottomrule
\end{tabular*}
\caption{Effect of DTC on DPO and DDO.}%
\label{tab:dtc-ddo-factorization}
\end{table}

\begin{table*}[t]
\centering
\scriptsize
\setlength{\tabcolsep}{2pt}
\renewcommand{\arraystretch}{0.96}
\resizebox{\textwidth}{!}{%
\begin{tabular}{l*{12}{c}}
\toprule
\multicolumn{1}{c}{Method} &
\multicolumn{4}{c}{Success rate $\uparrow$} &
\multicolumn{4}{c}{H-ESD $\uparrow$} &
\multicolumn{4}{c}{ESD $\uparrow$} \\
\cmidrule(lr){2-5} \cmidrule(lr){6-9} \cmidrule(lr){10-13}
& BabyAI & BabaIsAI & WebShop & Avg.
& BabyAI & BabaIsAI & WebShop & Avg.
& BabyAI & BabaIsAI & WebShop & Avg. \\
\midrule
Base        & 0.32 & 0.21 & 0.18 & 0.24 & 0.28 & 0.08 & 0.02 & 0.13 & 0.29 & 0.09 & 0.02 & 0.13 \\
Reference   & 0.82 & 0.90 & 0.26 & 0.66 & 0.41 & 0.37 & 0.15 & 0.31 & 0.49 & 0.43 & 0.17 & 0.36 \\
\midrule
DPO         & 0.84 & 0.94 & 0.26 & 0.68 & 0.36 & 0.31 & 0.11 & 0.26 & 0.43 & 0.38 & 0.12 & 0.31 \\
DivFreq     & 0.85 & 0.91 & 0.26 & 0.67 & 0.37 & 0.30 & 0.12 & 0.26 & 0.44 & 0.39 & 0.13 & 0.32 \\
DivProb     & 0.89 & 0.93 & 0.26 & 0.69 & 0.38 & 0.36 & 0.11 & 0.28 & 0.44 & 0.43 & 0.12 & 0.33 \\
TieDPO-RK   & 0.88 & 0.80 & 0.28 & 0.65 & 0.35 & 0.28 & 0.16 & 0.26 & 0.40 & 0.35 & 0.17 & 0.31 \\
TieDPO-Dav  & 0.86 & 0.77 & 0.16 & 0.60 & 0.38 & 0.30 & 0.13 & 0.27 & 0.43 & 0.38 & 0.13 & 0.31 \\
\midrule
DDO         & \textbf{0.92} & \textbf{0.99} & \textbf{0.36} & \textbf{0.76} & \textbf{0.45} & \textbf{0.38} & \textbf{0.21} & \textbf{0.35} & \textbf{0.52} & \textbf{0.47} & \textbf{0.22} & \textbf{0.40} \\
\bottomrule
\end{tabular}}
\caption{Main results across BabyAI, BabaIsAI, and WebShop. BabyAI and BabaIsAI report task averages, WebShop reports the shopping evaluation summary, and Avg. is the unweighted mean across benchmarks.}
\label{tab:main-benchmark-avg}
\end{table*}

These collection differences translate into post-training results.
\Cref{tab:dtc-ddo-factorization} compares DPO and DDO with and without DTC.
For DPO, adding DTC raises task success by \(8\) percentage points (pp) on BabyAI and \(16\) pp on BabaIsAI, although its effect on coverage is mixed on BabyAI.
For DDO, adding DTC raises task success by \(5\) pp on BabyAI and \(16\) pp on BabaIsAI.
Adding DTC also improves DDO's coverage metrics: on BabyAI, H-ESD and ESD each increase by \(0.06\); on BabaIsAI, they increase by \(0.10\) and \(0.12\), respectively.
Averaged across the two benchmarks, adding DTC to DDO improves task success by \(10.5\) pp, H-ESD by \(0.08\), and ESD by \(0.09\).

DTC broadens the collected set of successful trajectories and improves downstream post-training.
Under matched request budgets, DTC discovers more unique successful trajectories and provides branch supervision at decision states spanning early, middle, and late portions of the source trajectories.
The factorization in \Cref{tab:dtc-ddo-factorization} shows complementary contributions from DTC and RTO: DTC expands branch supervision and predominantly raises success, whereas RTO improves both success and coverage under either collection condition.
Combining DTC and RTO yields the strongest joint success and coverage result on both BabyAI and BabaIsAI.
We next fix the DTC resource and compare post-training objectives under the same state-aligned branch supervision.

\subsection{RTO Improves the Success--Coverage Frontier}
\label{sec:rto-results}

With DTC fixed, we compare post-training objectives on the same state-aligned branch sets. \Cref{tab:main-benchmark-avg} reports benchmark averages; full tables appear in Appendix~\ref{app:full-main-results}. The three benchmarks probe different aspects of successful strategy coverage---diverse successful action trajectories for the same instruction (BabyAI), diverse exploitable rule configurations (BabaIsAI), and diverse shopping behaviors ending in valid purchases (WebShop).

\paragraph{Joint success and coverage gains.}
DDO achieves the highest benchmark-average success (\(0.76\)), H-ESD (\(0.35\)), and ESD (\(0.40\)). Relative to the DPO row, these values correspond to gains of \(8\) pp, \(0.09\), and \(0.09\), respectively. The largest relative gain appears on WebShop, where H-ESD nearly doubles from \(0.11\) to \(0.21\) and ESD nearly doubles from \(0.12\) to \(0.22\), alongside a \(10\) pp success gain from \(0.26\) to \(0.36\).
Because WebShop trajectory classes incorporate purchase realization, the gain reflects broader successful strategy coverage under the composite class definition.

\paragraph{Coverage relative to task-SFT initialization.}
DDO raises task success by \(10\) pp and both coverage metrics by \(0.04\) over the shared Reference initialization. The remaining post-training methods stay below Reference on both coverage metrics; DPO, for example, raises average success from \(0.66\) to \(0.68\) while H-ESD and ESD each fall by \(0.05\).

\paragraph{Other post-training variants.}
Across the compared methods, DDO leads every benchmark-level aggregate and all three overall averages. DivFreq and DivProb, which filter pairs by frequency- or probability-based diversity, change success and coverage by at most \(\pm 0.02\) on average. TieDPO-RK and TieDPO-Dav reach lower average success than DPO (\(-3\) and \(-8\) pp), with the largest drops on BabaIsAI (\(-0.14\) and \(-0.17\) success) and, for TieDPO-Dav, WebShop (\(-0.10\)); their coverage matches DPO on average.

\section{Analysis}

\subsection{Strategic Recovery}

We use strategic recovery to test whether broader successful strategy coverage provides alternative routes after a local disruption.
For each successful source trajectory, we first sample one interior decision point.
We then replace the source action at that point with a different valid task action and roll out the same model from the edited prefix.
Because every source trajectory solves the task before the edit, the score measures recovery from the local action replacement rather than ordinary task success.
Appendix~\ref{app:expdetails} details the probe construction.

DDO achieves the highest recovery rate, \(75.2\%\), compared with \(70.1\%\) for DivFreq and \(69.7\%\) for DPO. After the initial trajectory is disrupted, DDO more often finds another viable trajectory to success, showing that its broader successful strategy coverage is accompanied by higher recovery.

\begin{figure}[t]
\centering
\includegraphics[width=\columnwidth]{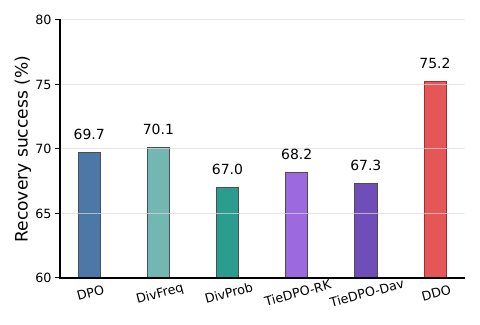}
\caption{Strategic recovery on BabyAI after local action replacement in successful source trajectories.}
\label{fig:babyai-steering-success}
\end{figure}

\subsection{Imitation Control}

To separate successful-branch exposure from same-state outcome comparisons and reference-relative targets, we train successful-only SFT models on the same successful DTC branches used by DDO.
We compare DDO with two imitation settings matched by optimizer steps and nominal epochs, respectively.
At matched optimizer steps, successful-only imitation reaches H-ESD \(0.39\) and ESD \(0.44\), compared with DDO's \(0.42\) and \(0.50\), while DDO retains a \(32\) pp advantage in task success.
At \(7.30\times\) exposure, imitation reaches H-ESD \(0.41\) and ESD \(0.47\), while DDO retains a \(24\) pp advantage in task success.
Successful-only SFT broadens coverage, while DDO retains a \(24\)--\(32\) pp advantage in task success under both matching conditions.

\subsection{Decoding Control}

We also test whether inference-time diversification can recover successful strategy coverage without post-training changes. We compare DDO against DPO under a range of sampling temperatures.

\begin{table}[t]
\centering
\small
\setlength{\tabcolsep}{3.0pt}
\renewcommand{\arraystretch}{0.98}
\begin{tabular*}{\columnwidth}{@{\extracolsep{\fill}}lcccc@{}}
\toprule
Method & Exposure & Success $\uparrow$ & H-ESD $\uparrow$ & ESD $\uparrow$ \\
\midrule
DDO, 5 ep. & \(1.00\times\) & 0.96 & 0.42 & 0.50 \\
\midrule
Imit. matched & \(1.00\times\) & 0.64 & 0.39 & 0.44 \\
Imit. 5 ep. & \(7.30\times\) & 0.72 & 0.41 & 0.47 \\
\bottomrule
\end{tabular*}
\caption{Successful-only imitation control averaged over BabyAI and BabaIsAI.}
\label{tab:imitation-control}
\end{table}

\begin{table}[t]
\centering
\small
\setlength{\tabcolsep}{2.0pt}
\renewcommand{\arraystretch}{0.96}
\begin{tabular*}{\columnwidth}{@{\extracolsep{\fill}}lcccc@{}}
\toprule
Method & \(T\) & Success $\uparrow$ & H-ESD $\uparrow$ & ESD $\uparrow$ \\
\midrule
DDO & 0.6 & 0.76 & 0.35 & 0.40 \\
\midrule
DPO + temp sweep & 0.0 & 0.60 & 0.22 & 0.25 \\
DPO + temp sweep & 0.1 & 0.62 & 0.20 & 0.24 \\
DPO + temp sweep & 0.6 & 0.68 & 0.26 & 0.31 \\
DPO + temp sweep & 1.0 & 0.66 & 0.29 & 0.35 \\
DPO + temp sweep & 1.5 & 0.67 & 0.33 & 0.39 \\
\bottomrule
\end{tabular*}
\caption{DPO temperature sweep averaged across BabyAI, BabaIsAI, and WebShop; DDO at \(T=0.6\) is shown for comparison.}
\label{tab:decode-compare}
\vspace{-4mm}
\end{table}
Simply increasing DPO's sampling temperature raises strategy coverage, but the gain comes with lower task success.
At \(T{=}1.5\), DPO reaches an ESD of \(0.39\), close to DDO's \(0.40\), while DDO retains a \(9\) pp advantage in task success.
Increasing temperature therefore recovers coverage by sacrificing success, whereas DDO improves the joint success and coverage result through post-training.

\section{Conclusion}

We present Direct Diversity Optimization (DDO), an offline post-training method for preserving multiple successful strategies in LLM agents trained from trajectory-level outcome labels.
DDO consists of two components: Divergence-Tree Collection (DTC) and the Reference-Relative Target-Odds Objective (RTO).
DTC builds state-aligned branch sets with per-branch outcome labels, and RTO trains the model toward reference-relative targets over successful branches.

Across all benchmarks, DDO achieves the strongest joint performance in task success and successful strategy coverage among the compared post-training methods.
DDO also achieves the highest recovery rate in the local action replacement evaluation.
Compared with successful-only imitation and decoding-time diversification, DDO achieves broader coverage together with higher task success by training on same-state outcome comparisons with reference-relative targets.
These results establish DDO as a training-time method that improves task success while broadening successful strategy coverage in sequential decision tasks.

\section*{Limitations}

For controlled benchmarking and method comparison, our experiments are confined to environments that support state reconstruction and the execution of alternative branches from a shared decision state under a common success predicate. This setting makes the outcomes of same-state alternative branches directly observable, but it excludes continuous control, partially observable or stochastic dynamics, multimodal observations, and real-world agents whose actions induce non-reversible external side effects. Extending DTC to settings without exact state reconstruction would require approximate state-aligned branch sets.
Such sets could be constructed from rollouts generated by learned simulators or world models, or from logged trajectories that contain comparable decision contexts and alternative outcomes.

\section*{Ethical Considerations}

\paragraph{License of Existing Assets.}
The models used in this work, Qwen3-1.7B and Qwen3.5-122B-A10B-FP8, are released under the Apache 2.0 license. The benchmark environments are used under their respective licenses: BabyAI under the BSD 3-Clause License, BabaIsAI under the MIT License, and WebShop under the MIT License with copyright attributed to Princeton Natural Language Processing. All assets and environments were used for academic, non-commercial evaluation purposes and in accordance with the applicable license terms.

\section*{Acknowledgments}

This work was supported by the Artificial Intelligence Industrial Convergence Cluster Development Project, funded by the Ministry of Science and ICT (MSIT), Korea, and Gwangju Metropolitan City.

\bibliography{custom}
\clearpage
\appendix

\section{Additional Results}

\subsection{Model-Scale Sensitivity}

We vary the expert and target models separately on BabyAI while keeping all other collection, training, and evaluation settings fixed (\Cref{tab:model-scale-sensitivity}).

\begin{table}[h]
\centering
\small
\setlength{\tabcolsep}{2.5pt}
\renewcommand{\arraystretch}{0.96}
\begin{tabular*}{\columnwidth}{@{\extracolsep{\fill}}llccc@{}}
\toprule
Scale & Policy & Success $\uparrow$ & H-ESD $\uparrow$ & ESD $\uparrow$ \\
\midrule
\multicolumn{5}{l}{\textit{Expert model scale}} \\
122B & DDO & 0.92 & 0.45 & 0.52 \\
35B  & DDO & 0.87 & 0.41 & 0.48 \\
\midrule
\multicolumn{5}{l}{\textit{Target model scale}} \\
1.7B & Reference & 0.82 & 0.41 & 0.49 \\
     & DDO       & 0.92 & 0.45 & 0.52 \\
4B   & Reference & 0.85 & 0.37 & 0.42 \\
     & DDO       & 0.91 & 0.40 & 0.45 \\
\bottomrule
\end{tabular*}
\caption{Model-scale sensitivity on BabyAI. Each block varies one model while holding the other fixed.}
\label{tab:model-scale-sensitivity}
\end{table}

Reducing the expert model size from 122B to 35B changes task success from \(0.92\) to \(0.87\), H-ESD from \(0.45\) to \(0.41\), and ESD from \(0.52\) to \(0.48\).
Even with the 35B expert model, DDO retains more than \(90\%\) of the corresponding 122B value for each metric: \(94.6\%\) for task success, \(91.1\%\) for H-ESD, and \(92.3\%\) for ESD.
The number of completed alternative branches also remains similar (\(3{,}329\) versus \(3{,}406\)).

DDO improves over its corresponding Reference at both target model sizes.
For the 1.7B target model, DDO improves task success, H-ESD, and ESD by \(0.10\), \(0.04\), and \(0.03\), respectively.
For the 4B target model, the corresponding gains are \(0.06\), \(0.03\), and \(0.03\).

\subsection{Coverage Growth}\label{app:coverage-growth}

\begin{figure}[h]
\centering
\includegraphics[width=\columnwidth]{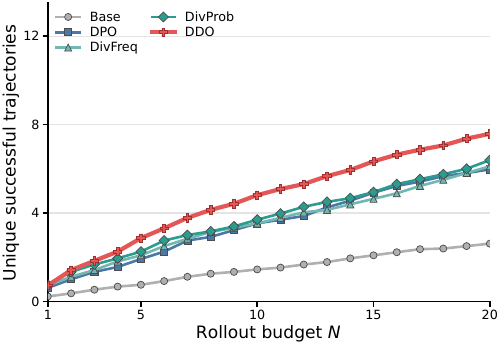}
\caption{Coverage growth over BabyAI, BabaIsAI, and WebShop. Each curve indicates the cumulative number of unique successful trajectories found within the first \(N\) rollouts.}
\label{fig:coverage-growth}
\end{figure}

Beyond final-budget scores, \Cref{fig:coverage-growth} traces the cumulative number of unique successful trajectories found within the first \(N\) rollouts on BabyAI, BabaIsAI, and WebShop. DDO stays above the other plotted methods across the sampled budget range, so its coverage gain is visible throughout sampling and at the final evaluation budget.

\subsection{Shared Cross-Task Post-Training}\label{app:shared-crosstask}

The main benchmark uses SFT and preference adapters for each task to provide controlled, matched comparisons. As an additional setting, we test whether DDO remains effective when both the initializer and post-training model are shared across tasks. On BabyAI, a single BabyAI-4 SFT adapter initializes all four tasks, and preference methods are trained on pooled BabyAI-4 preference data. The table reports BabyAI task averages under the same evaluation budgets and decoding settings as the main results.

\begin{table}[h]
\centering
\small
\setlength{\tabcolsep}{3.0pt}
\renewcommand{\arraystretch}{0.96}
\begin{tabular*}{\columnwidth}{@{\extracolsep{\fill}}lccc@{}}
\toprule
Method & Success $\uparrow$ & H-ESD $\uparrow$ & ESD $\uparrow$ \\
\midrule
SFT & 0.78 & 0.37 & 0.41 \\
DPO & 0.81 & 0.38 & 0.43 \\
DDO & \textbf{0.92} & \textbf{0.48} & \textbf{0.52} \\
\bottomrule
\end{tabular*}
\caption{Shared cross-task SFT and post-training on BabyAI using one BabyAI-4 SFT initialization and pooled preference data.}
\label{tab:crosstask-babyai}
\end{table}

\Cref{tab:crosstask-babyai} shows that DDO consistently improves both task success and successful strategy coverage in the shared-model setting. DDO achieves \(0.92\) success, \(0.48\) H-ESD, and \(0.52\) ESD, extending its joint success and coverage gains from task-specific adapters to pooled cross-task post-training.

\subsection{Branch-Set Target Distribution}
\label{sec:alpha-analysis}

In Eq.~\eqref{eq:q}, we interpret the RTO target as a log-odds interpolation between two endpoints: a uniform distribution on \(S^+\) (\(\alpha=0\), encoding within-success coverage) and the reference distribution restricted to \(S^+\) (\(\alpha=1\), preserving the relative ordering inherited from \(\pi_{\mathrm{ref}}\)).
The intermediate setting \(\alpha=0.5\) retains both signals.
\Cref{tab:ddo-alpha-sweep} compares these three regimes on BabyAI.

\begin{table}[h]
\centering
\small
\begin{tabular}{cccc}
\toprule
$\alpha$ & Success $\uparrow$ & H-ESD $\uparrow$ & ESD $\uparrow$ \\
\midrule
1.0 & 0.90 & 0.39 & 0.42 \\
0.5 & \textbf{0.92} & \textbf{0.45} & \textbf{0.52} \\
0.0 & 0.87 & 0.42 & 0.48 \\
\bottomrule
\end{tabular}
\caption{Effect of the target-distribution coefficient \(\alpha\) on BabyAI.}
\label{tab:ddo-alpha-sweep}
\end{table}

The intermediate target (\(\alpha=0.5\)) achieves the highest success (\(0.92\)), H-ESD (\(0.45\)), and ESD (\(0.52\)).
At the endpoints, \(\alpha=1\) preserves the reference-relative weighting over successful branches, while \(\alpha=0\) uses a uniform target.
The intermediate setting combines both target components and yields the strongest joint success and coverage result.

\subsection{Margin Sharpness Sensitivity}
\label{app:margin-sharpness}

The inverse-temperature parameter \(\beta\) controls the sharpness of the pairwise margin loss, determining how strongly deviations from the target pair odds are penalized.

\begin{figure}[h]
\centering
\includegraphics[width=\columnwidth]{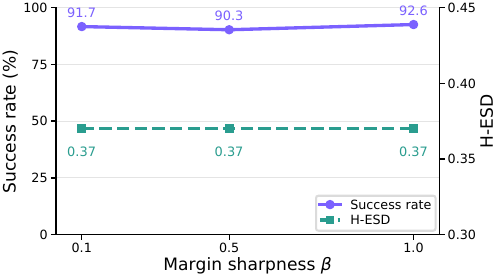}
\caption{Effect of the margin sharpness \(\beta\) on DDO, averaged over BabyAI and BabaIsAI. Axes report success rate and H-ESD.}
\label{fig:ddo-beta-sensitivity}
\end{figure}

\Cref{fig:ddo-beta-sensitivity} reports the DDO \(\beta\) sweep over \(\{0.1,0.5,1.0\}\).
Across this range, success changes only mildly, while H-ESD remains nearly unchanged after rounding.
DDO therefore varies little with margin sharpness across the tested values.

\subsection{Full Main Benchmark Tables}\label{app:full-main-results}

We provide the full benchmark results in \Cref{tab:babyai-main,tab:baba-main,tab:webshop-main}.
Bold marks the strongest preference-trained result in each column, including ties.

\paragraph{BabyAI.}
DDO leads on average success rate, H-ESD, and ESD. The task columns show the same pattern: success and H-ESD increase for Goto, Pick, Open, and Comp, and ESD increases on three tasks and reaches \(0.57\) on Comp. The Open and Comp columns make this pattern visible in tasks involving object interaction and multi-step composition; DDO improves both success and coverage there, so broader successful strategy coverage accompanies higher task success.

\paragraph{BabaIsAI.}
DDO is again the strongest method on average success rate, H-ESD, and ESD, with gains across Basic, Room, Stop, and Flex. The Stop column is particularly informative: DDO raises success to \(1.00\) while also improving both coverage metrics, indicating that the gains extend to rule manipulation tasks where successful behavior depends on changing the rule structure. DivProb and the tie-aware baselines are strongest or tied in a few individual coverage columns. DDO leads the benchmark averages while improving success and coverage together.

\paragraph{WebShop.}
DDO is the strongest method across all three reported metrics. Because WebShop counts unique successful trajectories using both trajectory structure and purchase realization, the gains reflect broader successful strategy coverage under the composite trajectory-class definition.

\begin{table*}[t]
\centering
\small
\setlength{\tabcolsep}{1.6pt}
\renewcommand{\arraystretch}{0.96}
\begin{tabular*}{\textwidth}{@{\extracolsep{\fill}}lccccc@{\hspace{1.0em}}ccccc@{\hspace{1.0em}}ccccc@{\hspace{0.6em}}}
\toprule
\multicolumn{1}{c}{Method} &
\multicolumn{5}{c}{Success rate $\uparrow$} &
\multicolumn{5}{c}{H-ESD $\uparrow$} &
\multicolumn{5}{c}{ESD $\uparrow$} \\
\cmidrule(lr){2-6} \cmidrule(lr){7-11} \cmidrule(lr){12-16}
            & Goto. & Pick. & Open. & Comp. & \textbf{Avg.}   & Goto. & Pick. & Open. & Comp. & \textbf{Avg.}   & Goto. & Pick. & Open. & Comp. & \textbf{Avg.} \\
\midrule
Base        & 0.76 & 0.22   & 0.04 & 0.26 & 0.32  & 0.50 & 0.28   & 0.02 & 0.30 & 0.28  & 0.55 & 0.28   & 0.02 & 0.30 & 0.29 \\
Reference   & 0.90 & 0.94   & 0.69 & 0.74 & 0.82  & 0.14 & 0.36   & 0.67 & 0.46 & 0.41  & 0.27 & 0.47   & 0.67 & 0.57 & 0.49 \\
\midrule
DPO         & 0.90 & 0.92   & 0.72 & 0.82 & 0.84  & 0.09 & 0.15   & 0.76 & 0.45 & 0.36  & 0.15 & 0.23   & 0.77 & \textbf{0.57} & 0.43 \\
DivFreq     & 0.94 & 0.96   & 0.70 & 0.80 & 0.85  & 0.13 & 0.19   & 0.75 & 0.40 & 0.37  & 0.18 & 0.30   & 0.75 & 0.52 & 0.44 \\
DivProb     & 0.94 & 0.98   & 0.78 & 0.86 & 0.89  & 0.13 & 0.16   & 0.79 & 0.45 & 0.38  & 0.18 & 0.22   & 0.80 & 0.55 & 0.44 \\
TieDPO-RK   & \textbf{1.00} & \textbf{1.00}   & 0.74 & 0.76 & 0.88  & 0.11 & 0.17   & 0.70 & 0.43 & 0.35  & 0.17 & 0.27   & 0.70 & 0.47 & 0.40 \\
TieDPO-Dav  & 0.94 & \textbf{1.00}   & 0.64 & 0.86 & 0.86  & 0.10 & \textbf{0.26}   & 0.72 & 0.45 & 0.38  & 0.18 & 0.33   & 0.72 & 0.50 & 0.43 \\
\midrule
DDO         & 0.94 & \textbf{1.00}   & \textbf{0.80} & \textbf{0.92} & \textbf{0.92}  & \textbf{0.21} & 0.24   & \textbf{0.84} & \textbf{0.51} & \textbf{0.45}  & \textbf{0.30} & \textbf{0.35}   & \textbf{0.85} & \textbf{0.57} & \textbf{0.52} \\
\bottomrule
\end{tabular*}
\caption{Main results on BabyAI. Avg. is the unweighted mean across the four tasks.}
\label{tab:babyai-main}
\end{table*}

\begin{table*}[t]
\centering
\small
\setlength{\tabcolsep}{1.6pt}
\renewcommand{\arraystretch}{0.96}
\begin{tabular*}{\textwidth}{@{\extracolsep{\fill}}lccccc@{\hspace{1.0em}}ccccc@{\hspace{1.0em}}ccccc@{\hspace{0.6em}}}
\toprule
\multicolumn{1}{c}{Method} &
\multicolumn{5}{c}{Success rate $\uparrow$} &
\multicolumn{5}{c}{H-ESD $\uparrow$} &
\multicolumn{5}{c}{ESD $\uparrow$} \\
\cmidrule(lr){2-6} \cmidrule(lr){7-11} \cmidrule(lr){12-16}
 & Basic. & Room. & Stop. & Flex. & \textbf{Avg.}
 & Basic. & Room. & Stop. & Flex. & \textbf{Avg.}
 & Basic. & Room. & Stop. & Flex. & \textbf{Avg.} \\
\midrule
Base        & 0.22 & 0.26 & 0.08 & 0.26 & 0.21  & 0.02 & 0.05 & 0.06 & 0.19 & 0.08    & 0.02 & 0.05 & 0.07 & 0.22 & 0.09 \\
Reference   & 0.94 & 0.97 & 0.86 & 0.84 & 0.90  & 0.20 & 0.11 & 0.59 & 0.59 & 0.37    & 0.29 & 0.17 & 0.62 & 0.64 & 0.43 \\
\midrule
DPO         & 0.96 & 0.92 & 0.92 & 0.96 & 0.94  & 0.10 & 0.06 & 0.58 & 0.50 & 0.31    & 0.19 & 0.10 & 0.65 & 0.59 & 0.38 \\
DivFreq     & 0.94 & 0.92 & 0.78 & \textbf{0.98} & 0.91  & 0.13 & \textbf{0.07} & 0.58 & 0.42 & 0.30    & 0.25 & 0.12 & 0.71 & 0.47 & 0.39 \\
DivProb     & 0.96 & 0.96 & 0.84 & 0.96 & 0.93  & 0.13 & 0.06 & 0.68 & \textbf{0.56} & 0.36    & 0.23 & 0.08 & 0.78 & \textbf{0.63} & 0.43 \\
TieDPO-RK   & 0.94 & 0.90 & 0.60 & 0.76 & 0.80  & 0.14 & \textbf{0.07} & 0.45 & 0.44 & 0.28    & 0.25 & 0.10 & 0.53 & 0.52 & 0.35 \\
TieDPO-Dav  & 0.90 & 0.94 & 0.50 & 0.74 & 0.77  & \textbf{0.19} & \textbf{0.07} & 0.42 & 0.54 & 0.30    & 0.28 & 0.10 & 0.50 & 0.62 & 0.38 \\
\midrule
DDO         & \textbf{0.98} & \textbf{0.98} & \textbf{1.00} & \textbf{0.98} & \textbf{0.99}  & 0.18 & \textbf{0.07} & \textbf{0.75} & 0.52 & \textbf{0.38}    & \textbf{0.30} & \textbf{0.13} & \textbf{0.81} & \textbf{0.63} & \textbf{0.47} \\
\bottomrule
\end{tabular*}
\caption{Main results on BabaIsAI. Avg. is the unweighted mean across the four tasks.}
\label{tab:baba-main}
\end{table*}

\begin{table}[t]
\centering
\small
\setlength{\tabcolsep}{3.0pt}
\renewcommand{\arraystretch}{0.96}
\begin{tabular*}{\columnwidth}{@{\extracolsep{\fill}}lccc@{}}
\toprule
Method & Success rate $\uparrow$ & H-ESD $\uparrow$ & ESD $\uparrow$ \\
\midrule
Base        & 0.18 & 0.02 & 0.02 \\
Reference   & 0.26 & 0.15 & 0.17 \\
\midrule
DPO         & 0.26 & 0.11 & 0.12 \\
DivFreq     & 0.26 & 0.12 & 0.13 \\
DivProb     & 0.26 & 0.11 & 0.12 \\
TieDPO-RK   & 0.28 & 0.16 & 0.17 \\
TieDPO-Dav  & 0.16 & 0.13 & 0.13 \\
\midrule
DDO         & \textbf{0.36} & \textbf{0.21} & \textbf{0.22} \\
\bottomrule
\end{tabular*}
\caption{Main results on WebShop.}
\label{tab:webshop-main}
\end{table}

\section{Additional Method Details}
\label{app:md}

\subsection{Action Parsing and Branch Output Probabilities}
\label{app:iface}

At each environment step, the LLM emits a model output \(y_t\), and a benchmark-specific parser extracts the executable action for the environment.
Some prompts structure \(y_t\) as reasoning text followed by an action field; DDO assigns sequence probability to the complete model output and uses the parsed action for execution.
For a DTC record \(C=(x,B,\{(y_j,r_j)\}_{j\in B})\), \(\pi_\theta(y_j\mid x)\) is the sequence probability of the model output under standard left-to-right token factorization.
The rollout determines the branch outcome label \(r_j\).

\subsection{DTC Collection Details}
\label{app:dtc}

The helper routines in \Cref{alg:dtc} are instantiated as follows.
Across benchmarks, we set \(K_d\) to at most \(5\) and \(K_a\) to at most \(3\); shorter trajectories and branch rejection can yield fewer retained points or alternatives.
\(\textsc{Select}\) allocates the divergence budget \(K_d\) over interior decision points \(1,\ldots,T-1\) so that selected steps cover different trajectory depths.
\(\textsc{Restore}\) returns the decision state associated with the selected source prefix, or \(\bot\) when the state cannot be constructed under the benchmark environment.
\(\textsc{Alt}\) queries the expert model \(E\) for at most \(K_a\) alternative outputs from the shared state \(x\), required to differ from the source output under the benchmark's executable-action normalization.
\(\textsc{Reject}\) removes outputs that cannot be executed or that do not provide a distinct executable branch decision, and retained outputs are rolled out by \(E\) to obtain outcome labels.

\section{Experimental Details}\label{app:expdetails}

\subsection{Benchmark Details}\label{app:benchmark-details}
BabyAI provides controlled navigation and object interaction tasks with executable text actions. We use four tasks: \textit{goto}, \textit{pickup}, \textit{open}, and \textit{pick-up sequence go-to}. They are reported in \Cref{tab:babyai-main} as Goto, Pick, Open, and Comp. For trajectory-class normalization, we collapse modulo-four same-direction turn repetitions and remove alternating turn blocks that cancel out, both of which leave the agent's pose unchanged.

BabaIsAI provides tasks based on rule manipulation with executable text actions. We use four tasks: \textit{goto}, \textit{two-room goto}, \textit{two-room break-stop goto}, and \textit{two-room optional break-stop goto}. They are reported in \Cref{tab:baba-main} as Basic, Room, Stop, and Flex. For trajectory-class normalization, we drop non-terminal steps with no observation change and two-step inverse moves that return the agent to a prior state, both of which leave the environment state unchanged.

WebShop provides shopping tasks in which an agent searches, inspects products, selects options, and purchases an item for a user instruction. For trajectory-class normalization, the class combines the purchased item, selected options, and normalized trajectory structure, making coverage invariant to surface variation in the action trace.
Figure~\ref{fig:webshop-trajectory-class-examples} shows how this rule merges query variants along the same route while distinguishing different search and navigation routes to the same purchase.

\begin{figure}[t]
\begin{tcolorbox}[width=\linewidth,colback=white,colframe=black]
\small
\textbf{Same class (query variation).} Two successful rollouts begin with \texttt{search[blue coated steel end table]} and \texttt{search[blue-coated steel end table]}, follow the same normalized trajectory structure, and purchase item \texttt{B08MF23ZPL} with the \texttt{blue} option.

\textbf{Different classes (route variation).} Two successful DDO rollouts purchase item \texttt{B09P572DP9} with the \texttt{redblack} option. Their normalized routes are search \(\rightarrow\) product inspection \(\rightarrow\) refined search \(\rightarrow\) purchase and search \(\rightarrow\) pagination \(\rightarrow\) refined search \(\rightarrow\) purchase.
\end{tcolorbox}
\caption{WebShop trajectory-class examples.}
\label{fig:webshop-trajectory-class-examples}
\end{figure}

Example input prompts and LLM outputs for all three benchmarks are shown in Appendix~\ref{app:example-llm-calls}.

\subsection{Models and Optimization}

For the experiments, expert trajectories are collected with \mbox{Qwen3.5-122B-A10B-FP8} under the shared thought-action format, and all optimized models use \mbox{Qwen3-1.7B}~\citep{yang2025qwen3,qwen2026qwen35}. The FP8 model provides the expert trajectory source. SFT and preference post-training for the target model use bf16 mixed precision. Base denotes the unadapted target model before SFT or preference optimization for each task. For each benchmark task, the Reference model is a shared SFT adapter trained from the base model for 10 epochs. DPO \cite{rafailov2023direct}, DivFreq, DivProb, TieDPO-RK, TieDPO-Dav, and DDO all start from this same Reference adapter, which is also used as the frozen reference model.

LoRA~\citep{hu2022lora} is applied to the Q/K/V/O attention projections and MLP projections, with rank \(32\), alpha \(32\), and dropout \(0.05\). All preference runs use AdamW~\citep{loshchilov2019decoupled} with betas \((0.9,0.999)\), epsilon \(10^{-8}\), weight decay \(0.0\), gradient clipping at norm \(1.0\), and a linear learning-rate schedule. Unless otherwise stated, preference-trained methods use learning rate \(5\times 10^{-6}\), warmup ratio \(0.03\), and \(\beta=0.1\). DDO uses \(\alpha=0.5\). To construct \(q_{\mathrm{ref}}\), the reference model scores the executable action field of each branch output.

\subsection{DTC Collection Cost}
\label{app:end-to-end-cost}

\begin{table}[h]
\centering
\small
\renewcommand{\arraystretch}{0.96}
\setlength{\tabcolsep}{3.0pt}
\begin{tabular*}{\columnwidth}{@{\extracolsep{\fill}}lrrr@{}}
\toprule
Collector & Requests & Tokens & Env. steps \\
\midrule
DTC & 28.6 & 59.7K & 48.3 \\
Without DTC & 65.3 & 156.0K & 64.7 \\
\bottomrule
\end{tabular*}
\caption{Collection cost per unique successful trajectory on BabyAI and BabaIsAI.}
\label{tab:end-to-end-cost}
\end{table}

On BabyAI and BabaIsAI, DTC uses 44\% of the expert requests, 38\% of the expert tokens, and 75\% of the environment steps used by Without DTC per unique successful trajectory. RTO requires no additional expert calls or environment interactions.

\subsection{Evaluation and Ablation Protocols}

A fixed seed protocol is used throughout. Training collection and evaluation use disjoint task seeds or WebShop evaluation sessions. BabyAI and BabaIsAI use the epoch-5 preference policies, while WebShop uses the epoch-15 preference policies. 
Main BabyAI and BabaIsAI results use sampled decoding with a temperature of \(0.6\), top-\(p\) of \(0.95\), and a max token limit of \(8192\).
WebShop coverage evaluation uses the same model and decoding/environment configuration as the corresponding success-rate evaluation.

\paragraph{Main results.}
For BabyAI and BabaIsAI, success rate is computed from 50 rollouts per evaluation item. H-ESD and ESD use 20 rollouts for each of three seeds used for coverage evaluation and are averaged. For WebShop, success rate is reported over 50 evaluation sessions. WebShop coverage uses sessions \(500\), \(501\), and \(502\), with 20 rollouts per session and varied LLM sampling seeds. Failed purchases, invalid actions, retry exhaustion, and max-step failures count toward the fixed rollout denominator. Rollouts interrupted by system failures or missing model outputs are excluded and replaced with reruns.

\paragraph{Component comparison.}
The component comparison in \Cref{tab:dtc-ddo-factorization} uses the BabyAI and BabaIsAI evaluation protocol described above. Settings without DTC train on same-state comparisons derived from separately sampled rollouts: successful and failed trajectories are sampled for the same task seed, converted into a preference comparison at their first shared decision state divergence, and matched to the corresponding budget for training comparisons. Settings with DTC use the state-aligned branch sets. All four settings use the corresponding DPO and DDO hyperparameters in this appendix and report benchmark averages.

\paragraph{Strategic recovery.}
The strategic recovery probe in \Cref{fig:babyai-steering-success} is run on BabyAI. For each method-task pair, we use successful evaluation rollouts, sample one interior decision point per rollout, replace the source action with a different valid task action, and roll out the same model from the edited prefix. To keep the probe balanced, the probe uses up to 30 successful rollouts per method-task pair. The reported value is success over valid edited rollouts.

\paragraph{Imitation control.}
The imitation control in \Cref{tab:imitation-control} is run on BabyAI and BabaIsAI. It starts from the same SFT initialization for each task and continues SFT on all successful DTC branches. The matched condition uses the same optimizer step count as DDO. The full-epoch condition runs successful-only imitation for the same nominal epoch count as DDO. Exposure is the number of processed training examples normalized by the DDO comparison exposure. Metrics use the two-benchmark aggregate.

\paragraph{Decoding controls.}
The decoding diversification ablation in \Cref{tab:decode-compare} reports an equal-weight benchmark macro average over BabyAI, BabaIsAI, and WebShop. The temperature sweep evaluates DPO at \(T\in\{0.0,0.1,0.6,1.0,1.5\}\) with one rollout per item. The DDO setting uses sampled decoding at \(T=0.6\). All preference methods use epoch-5 policies for BabyAI and BabaIsAI and epoch-15 policies for WebShop.

\paragraph{Coverage growth.}
Coverage growth curves use the BabyAI, BabaIsAI, and WebShop rollouts used for coverage evaluation. For each \(N\), task averages are computed within each benchmark and then summarized with the three-benchmark aggregate.

\paragraph{Branch-Set Target Distribution.}
The target distribution comparison in \Cref{tab:ddo-alpha-sweep} uses \(\beta=0.1\).
Under the parameterization in Eq.~\eqref{eq:q}, \(\alpha=0.0\) targets a uniform distribution over the observed successful branches, \(\alpha=0.5\) contracts the reference log-odds among successful
branches halfway toward uniformity, and \(\alpha=1.0\) preserves the reference distribution restricted to successful branches.

\paragraph{Margin sharpness sensitivity.}
The \(\beta\) sensitivity figure fixes \(\alpha=0.5\) and varies \(\beta\), using the two-benchmark aggregate.

\section{Example LLM Calls}
\label{app:example-llm-calls}

This appendix shows representative input prompts and LLM outputs for each
benchmark used in our evaluation.
The three benchmarks share a common thought-action response format: the model
produces a reasoning trace followed by an executable action, which a
benchmark-specific parser extracts and forwards to the environment
(Appendix~\ref{app:iface}).
Figures~\ref{fig:babyai-prompt-output}--\ref{fig:webshop-prompt-output}
illustrate this format on a single decision step in each benchmark.

\paragraph{BabyAI.}
The input prompt in Figure~\ref{fig:babyai-prompt-output} contains the task
instruction, a textual rendering of the gridworld observation around the agent,
the recent interaction history, and the admissible action set.
The LLM output reasons over the visible objects and the current goal before
emitting a single low-level navigation or manipulation action.

\paragraph{BabaIsAI.}
The input prompt in Figure~\ref{fig:babaisyou-prompt-output} additionally
exposes the active rule configuration of the puzzle, since success depends on
identifying and, when necessary, manipulating the rules.
The model reasons over the grid state and active rules, then outputs either a
goal-directed action or an action that changes the puzzle's rule configuration.

\paragraph{WebShop.}
The input prompt in Figure~\ref{fig:webshop-prompt-output} provides the user's
purchase instruction, the contents of the current page, and the available
interaction options such as search, product clicks, option selection, and
purchase.
The LLM output reasons about product attributes relative to the instruction
before emitting a single page interaction.

\clearpage

\begin{figure*}[p]
  \centering
  \begin{tcolorbox}[
    width=0.98\textwidth,
    colback=white,
    colframe=black,
    title=BabyAI,
    fonttitle=\bfseries,
  ]
  \textbf{Input prompt.}
\begin{lstlisting}[style=ddoexample]
[user message 1]
You are an agent playing a simple navigation game. Your goal is to go to the green box. The following are the possible actions you can take in the game, followed by a short description of each action:

turn left: turn to the left,
turn right: turn to the right,
go forward: take one step forward,
pick up: pick up the object directly in front of you (1 step forward),
drop: place the object you are carrying on an empty tile directly in front of you (1 step forward),
toggle: interact with the object directly in front of you; effects depend on object and state (1 step forward).

In a moment I will present you an observation.

Tips:
- Use 'pick up' to collect carryable items directly in front of you (1 step forward).
- Use 'toggle' to interact with the object directly in front of you.
- Use 'drop' to place the carried object on an empty tile directly in front of you (1 step forward).
- Some actions may have no effect if their preconditions are not met (for example, blocked front tile or no carried object).
- Avoid repeating the same action over and over if the observation doesn't change.

PLAY!

[user message 2]
Current Observation:
a wall 6 steps forward
a wall 3 steps left
a green box 1 step right and 5 steps forward

Your response should use the following format:

Thought: <your thoughts>
Action: <your next action>
\end{lstlisting}
  \textbf{LLM output.}
\begin{lstlisting}[style=ddoexample]
Thought: I need to reach the green box. The green box is 1 step right and 5 steps forward. But there is a wall 6 steps forward and a wall 3 steps left. I should first move forward to avoid the walls. Let's try moving forward.

Action: go forward
\end{lstlisting}
  \end{tcolorbox}
  \caption{Example BabyAI input prompt and LLM output.}
  \label{fig:babyai-prompt-output}
\end{figure*}

\begin{figure*}[p]
  \centering
  \begin{tcolorbox}[
    width=0.98\textwidth,
    colback=white,
    colframe=black,
    title=BabaIsAI,
    fonttitle=\bfseries,
  ]
  \textbf{Input prompt: task instructions.}
\begin{lstlisting}[style=ddoexample,firstline=1,lastline=20]
[user message 1]
BabaIsAI is a puzzle game where you can manipulate the rules of each level. The following are the possible actions you can take in the game, followed by a short description of each action:

idle: wait for one step,
up: take one step up,
right: take one step to the right,
down: take one step down,
left: take one step to the left.

Tips:
- Examine the level carefully, noting all objects and text blocks present.
- Identify the current rules, which are formed by text blocks in the format "[Subject] IS [Property]" (e.g. "BABA IS YOU").
- Consider how you can change or create new rules by moving text blocks around.
- Remember that you can only move objects or text that are not defined as "STOP" or similar immovable properties.
- Your goal is usually to reach an object defined as "WIN", but this can be changed.
- Think creatively about how changing rules can alter the properties and behaviors of objects in unexpected ways.
- If stuck, try breaking apart existing rules or forming completely new ones.
- Sometimes the solution involves making yourself a different object or changing what counts as the win condition.

PLAY!
\end{lstlisting}
  \end{tcolorbox}
  \caption{Example BabaIsAI input prompt (part 1).}
  \label{fig:babaisyou-prompt-output}
\end{figure*}

\begin{figure*}[p]
  \ContinuedFloat
  \centering
  \begin{tcolorbox}[
    width=0.98\textwidth,
    colback=white,
    colframe=black,
    title=BabaIsAI (continued),
    fonttitle=\bfseries,
  ]
  \textbf{Input prompt: current state and output format.}
\begin{lstlisting}[style=ddoexample,firstline=22,lastline=57]
[user message 2]
Current Observation:
Active rules:
ball is win
wall is stop
baba is you

Objects on the map:
wall 4 steps to the right and 4 step up
rule `ball` 7 steps to the right and 4 step up
rule `is` 8 steps to the right and 4 step up
rule `win` 9 steps to the right and 4 step up
rule `wall` 1 step to the left and 3 step up
rule `is` 3 step up
rule `stop` 1 step to the right and 3 step up
wall 4 steps to the right and 3 step up
ball 5 steps to the right and 3 step up
wall 4 steps to the right and 2 step up
wall 4 steps to the right and 1 step up
wall 4 steps to the right
wall 4 steps to the right and 1 step down
rule `baba` 1 step to the left and 2 steps down
rule `is` 2 steps down
rule `you` 1 step to the right and 2 steps down
wall 4 steps to the right and 2 steps down

Your response should use the following format:

Thought: <one short thought>
Action: <exactly one of: idle, up, right, down, left>

Rules:
- The Action line must be exactly one of: idle, up, right, down, left.
- Do not add any other words, punctuation, or explanation on the Action line.
- Output exactly one Thought line and exactly one Action line.
- Do not output anything after the Action line.
\end{lstlisting}
  \textbf{LLM output.}
\begin{lstlisting}[style=ddoexample]
Thought: I need to break the "WALL IS STOP" rule to pass through the wall barrier blocking the path to the ball.
Action: left
\end{lstlisting}
  \end{tcolorbox}
  \caption{Example BabaIsAI input prompt and LLM output (part 2).}
\end{figure*}

\begin{figure*}[p]
  \centering
  \begin{tcolorbox}[
    width=0.98\textwidth,
    colback=white,
    colframe=black,
    title=WebShop,
    fonttitle=\bfseries,
  ]
  \textbf{Input prompt.}
\begin{lstlisting}[style=ddoexample]
[user message 1]
You are a sequential WebShop text-environment agent.
Your mission is to satisfy the shopping instruction by navigating pages,
selecting required options, and buying the best matching product.

Valid action forms:
- search[keywords]: search from the search page only.
- click[value]: click a visible button, product id, navigation control, option,
  or click[buy now].

Choose exactly one valid action at each decision step.

[user message 2]
Current Observation:
Instruction: [SEP] i am looking for oral hygiene dental tools of design: set of 4, and price lower than 40.00 dollars [SEP] Back to Search [SEP] < Prev [SEP] design [SEP] dental curette [SEP] dental probe [SEP] mouth mirror [SEP] set of 4 (dental hygiene kit) [SEP] tarter scraper [SEP] Oral Care Dental Tools, Professional Dental Picks for Home use, Dental Tarter Scraper for Tooth Stains Remover, Plaque Remover for Teeth, Dentist Pick Stainless Steel (Dental Probe) [SEP] Price: $4.99 [SEP] Rating: N.A. [SEP] Description [SEP] Features [SEP] Reviews [SEP] Buy Now

Available actions for this state:
- click[back to search]
- click[< prev]
- click[description]
- click[features]
- click[reviews]
- click[buy now]
- click[dental curette]
- click[dental probe]
- click[mouth mirror]
- click[set of 4 (dental hygiene kit)]
- click[tarter scraper]

Output exactly two lines and nothing else:

Thought: <your thoughts>
Action: <one valid action>
\end{lstlisting}
  \textbf{LLM output.}
\begin{lstlisting}[style=ddoexample]
Thought: The current item is a dental probe, not a set of 4. I need to select the set option.
Action: click[set of 4 (dental hygiene kit)]
\end{lstlisting}
  \end{tcolorbox}
  \caption{Example WebShop input prompt and LLM output.}
  \label{fig:webshop-prompt-output}
\end{figure*}

\end{document}